\documentclass[nohyperref]{article}

\usepackage{microtype}
\usepackage{graphicx}
\usepackage{subfigure}
\usepackage{booktabs} % for professional tables

\usepackage{hyperref}

\usepackage[accepted]{icml2022}

\usepackage{amsmath}
\usepackage{amssymb}
\usepackage{mathtools}
\usepackage{amsthm}
\usepackage{bbm}
\usepackage{color}

\usepackage[capitalize,noabbrev]{cleveref}

\usepackage[ruled,vlined]{algorithm2e}
\usepackage{multirow}

\theoremstyle{plain}

\theoremstyle{definition}

\theoremstyle{remark}

\usepackage[textsize=tiny]{todonotes}

\icmltitlerunning{PAC-Net: A Model Pruning Approach to Inductive Transfer Learning}

\begin{document}

\twocolumn[
\icmltitle{PAC-Net: A Model Pruning Approach to Inductive Transfer Learning}



\icmlsetsymbol{equal}{*}

\begin{icmlauthorlist}
\icmlauthor{Sanghoon Myung}{C}
\icmlauthor{In Huh}{C}
\icmlauthor{Wonik Jang}{C}
\icmlauthor{Jae Myung Choe}{C}
\\
\icmlauthor{Jisu Ryu}{C}
\icmlauthor{Dae Sin Kim}{C}
\icmlauthor{Kee-Eung Kim}{K}
\icmlauthor{Changwook Jeong}{U}
\end{icmlauthorlist}

\icmlaffiliation{C}{CSE Team, Innovation Center, Samsung Electronics} %
\icmlaffiliation{U}{Graduate School of Semicondutor Materials and Devices Engineering, UNIST} 
\icmlaffiliation{K}{Kim Jaechul Graduate School of AI, KAIST}

\icmlcorrespondingauthor{Sanghoon Myung}{shoon.myung@samsung.com}
\icmlcorrespondingauthor{Changwook Jeong}{changwook.jeong@unist.ac.kr}

\icmlkeywords{Transfer Learning, Pruning, Partial Differential Equations, Ordinary Differential Equations, Catastrophic Forgetting}

\vskip 0.3in
]



\printAffiliationsAndNotice{}  % leave blank if no need to mention equal contribution

\begin{abstract}

Inductive transfer learning aims to learn from a small amount of training data for the target task by utilizing a pre-trained model from the source task. Most strategies that involve large-scale deep learning models adopt initialization with the pre-trained model and fine-tuning for the target task. However, when using over-parameterized models, we can often prune the model without sacrificing the accuracy of the source task. This motivates us to adopt model pruning for transfer learning with deep learning models. In this paper, we propose PAC-Net, a simple yet effective approach for transfer learning based on pruning. PAC-Net consists of three steps: Prune, Allocate, and Calibrate (PAC). The main idea behind these steps is to identify essential weights for the source task, fine-tune on the source task by updating the essential weights, and then calibrate on the target task by updating the remaining redundant weights. Under the various and extensive set of inductive transfer learning experiments, we show that our method achieves state-of-the-art performance by a large margin.

\end{abstract}

\section{Introduction}
\label{sec1}
Deep neural networks trained with massive data have exceeded humans in accuracy across various tasks, e.g., vision recognition~\cite{resnet} or natural language processing~\cite{squad}. However, in real-world situations where the training samples are scarce, the performance of deep neural networks often deteriorates significantly. Transfer learning techniques have been introduced to overcome such performance degradation in the small data regime \cite{pan2009survey}. In particular, inductive transfer learning is a promising framework that allows learning with a limited target dataset by leveraging a pre-trained source model, under the assumption that the source and target tasks are closely related to each other \cite{xuhong2018explicit} \cite{li2019delta}. For successful inductive transfer learning with deep neural networks, it is crucial to utilize the weights of the source model by transferring the core knowledge embedded in the weights to the target model. Typically, it is realized by fine-tuning \cite{FT} which initializes the weight vector of the target model as that pre-trained on the source task. Some follow-up studies \cite{xuhong2018explicit, li2021improved, li2019delta} further regularize the target model to prevent diverging from the source weight vector during the fine-tuning. 

\begin{figure*}[t]
\begin{center}
\end{center}
   \centering
   \includegraphics[width=1\textwidth]{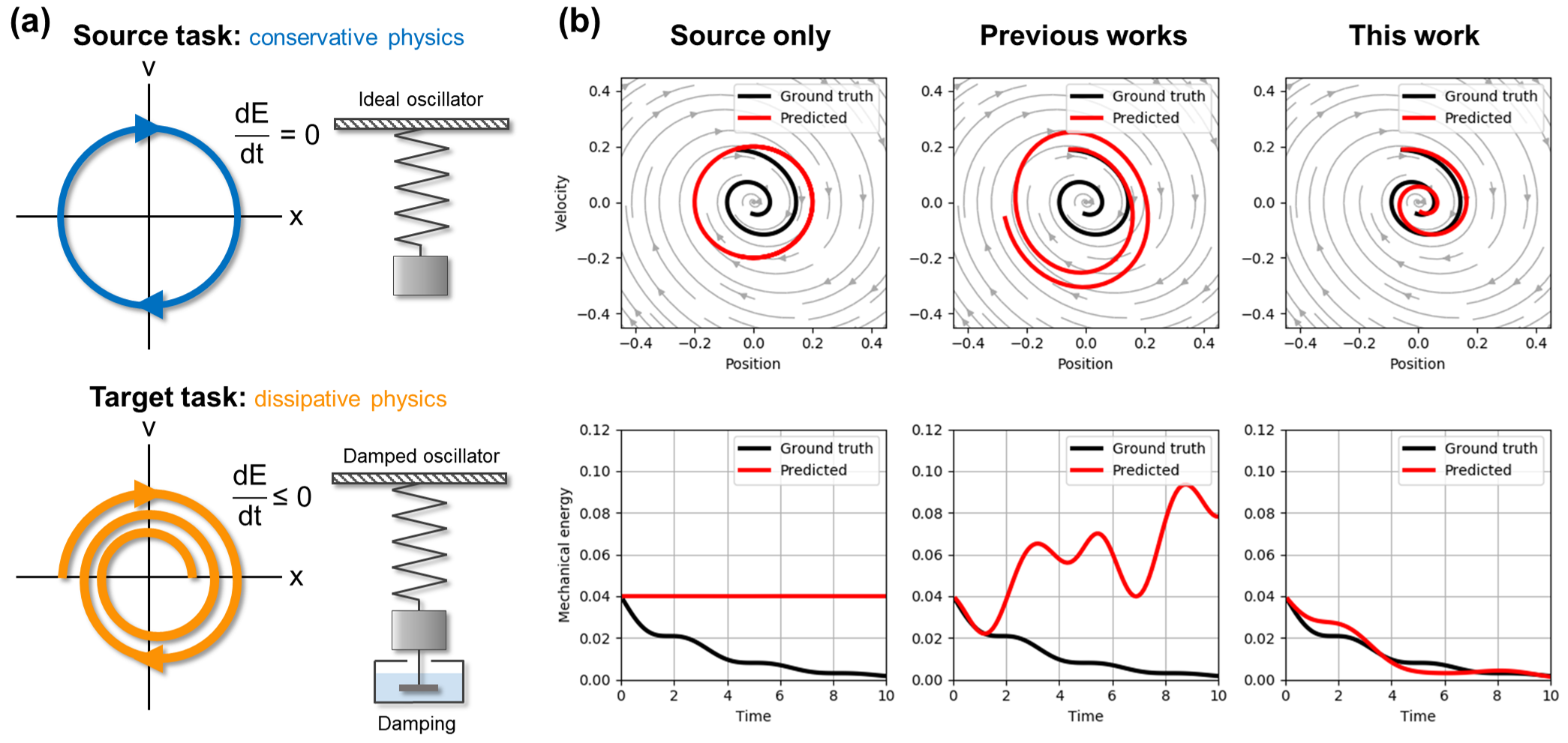}

   \caption{\textbf{(a) Description of the source and target tasks. (b) Phase and energy-versus-time plot.} What (a) and the leftmost of (b) illustrate is that the source and target systems are conservative and dissipative, respectively. What stands out in the rightmost of (b) is that our work can predict real physics by keeping the source knowledge (conservation of energy) and learning the target data (loss of energy) while the center (previous works) of (b) cannot.}
   \vskip -0.2in

\label{fig:1}
\end{figure*}

However, we found that these methods can easily overfit the target data despite their novel initialization and regularization. We hypothesize that this is because the model cannot preserve the core source knowledge when learning the target data. Consider a simple problem of learning dynamics for a real-world oscillator (e.g., a spring-mass-damper system), as depicted in Figure \ref{fig:1}. The conservation of energy is one of the principal laws that govern the dynamical behavior of the ideal spring-mass system (top of Figure \ref{fig:1} (a)). On the other hand, the real-world oscillator interacts with the environment, thus losing its mechanical energy due to the damping factor (bottom of Figure \ref{fig:1} (a)). What we expect from inductive transfer learning is that the model learns the law of the conservation of energy from the source task, then recognizes the damping term that causes the energy loss from a small number of measured target samples. The leftmost column of Figure \ref{fig:1} (b) shows the model pre-trained on the source data has learned the energy conservation principle. After the fine-tuning process \cite{FT, xuhong2018explicit, li2019delta}, unfortunately, the model overfits on few samples and forgets the law of conservation of energy; as shown in the center of Figure \ref{fig:1} (b), the transferred model predicts the oscillator gains its energy from the environment, which violates the fundamental law of physics. 

To overcome this problem, we propose a simple yet effective algorithm that can learn the target task (e.g., loss of energy) while preserving the relevant knowledge obtained from the source task (e.g., conservation of energy), as shown in the rightmost of Figure \ref{fig:1} (b). This is achieved by decomposing the weights into two disjoint sets, where one is in charge of learning the source task and the other for the target task, by adopting the pruning technique. More specifically, the algorithm consists of the following three steps: pruning, allocation, and calibration (PAC), which we call PAC-Net. In the pruning step, we sparsify the deep neural network pre-trained on the source task dataset. Then, we allocate the unpruned ($\mathbf{w}_U$) and pruned weights ($\mathbf{w}_P$) to transferable and calibratable parts, respectively. Finally, we calibrate $\mathbf{w}_P$ with the target task samples.
The contribution of our work is summarized below:
\begin{itemize}
  \item Our method provides a unified transfer learning approach for classification and regression as well as non-convolutional neural networks, which is of more general applicability compared to previous work.
  \item To our best knowledge, our paper is the first study to provide transfer learning experiments on solving ODEs and PDEs, an impactful task for machine learning to solve scientific problems.
  \item Our method for preserving source knowledge through pruning with regularization makes the model mitigate catastrophic forgetting, which achieves state-of-the-art performance.
\end{itemize}

This paper is organized as follows. Section \ref{sec2} reviews the previous works related to our work. Section \ref{sec3} describes PAC-Net in detail. Section \ref{sec4} shows experimental results of the proposed method on various domains. Section \ref{sec5} concludes this paper with a brief remark.

\section{Related Works} \label{sec2}

\begin{figure*}[t!]
   \centering
   \includegraphics[width=1\textwidth]{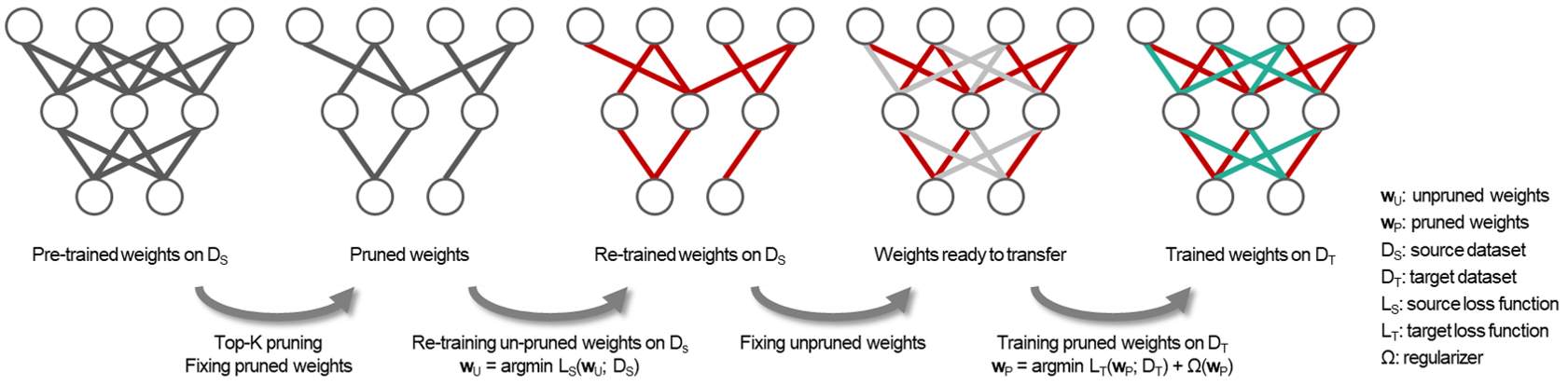}

   \caption{
   PAC-Net operates in three steps. In the first step, PAC-Net prunes inconspicuous weights after initially training the network with source data. In the second step, PAC-Net allocates the unpruned and pruned weights for the source and the target task, respectively. in this step, PAC-Net re-trains the source task with unpruned weights and keeps the pruned weights to zero. At last, PAC-Net calibrates the pruned weights with few target samples while regularizing them to be not too far from zero.}
   \vskip -0.2in

\label{fig:2}
\end{figure*}

\subsection{Transfer Learning} \label{sec2.1}
Following the nomenclature in \cite{pan2009survey}, transfer learning can be categorized
into transductive transfer learning (i.e. domain adaptation) and inductive transfer learning. We briefly review some of the previous work on domain adaptation and inductive transfer learning. 

\paragraph{Domain adaptation.} Transductive transfer learning is usually called domain adaptation, where tasks are the same and the domains are different but related to each other. The domain adaptation is categorized into supervised (with a few labels in the target domain) and unsupervised (without labels in the target domain) settings. In supervised domain adaptation, classification and contrast semantic alignment (CCSA)~\cite{motiian2017unified} and domain adaptation using stochastic neighborhood embedding \cite{dsne} use two stream architectures and 
introduce loss functions to reduce the distances among intra-class data pairs in the embedding space while maximizing the distances
among inter-class data pairs. In unsupervised domain adaptation, many approaches attempt to align statistical distribution shift between the source and target domains \cite{coral, mmd, zellinger2017central}, or employ discriminators for domain alignment \cite{adda, dann}.

\paragraph{Inductive transfer learning.} On the other hand, inductive transfer learning, which is the focus of this paper, refers to the case where tasks are different but related to each other, while domains are the same. Typically, the relationship between learning simulation models (the source) and real-world physical phenomena (the target) can be explained by inductive transfer learning task. It is crucial how to utilize the weights of the source task when training on the target task. Here, let us briefly review the literature. 

Boosting ensemble employs additional networks to correct errors made by networks learning the source task. AdaBoost \cite{adaboost} is an adaptive method in which subsequent weak learners train instances that are misclassified by previous classifiers. \cite{tradaboost} and \cite{boostingforreg} extended AdaBoost to transfer learning for classification and regression. These approaches, however, incur high computational costs when employing a large network as the base learner.

Another approach utilizes the weights of the source model as initial weights to train the target task, known as fine-tuning. \cite{FT} has demonstrated the representation, learned from the source task, can be transferred to the target tasks. \cite{xuhong2018explicit} compares a number of different regularization strategies that make the weights of the source model and the target model similar during fine-tuning. \cite{li2019delta} preserves a subset of feature maps from the source model with channel-wise attention to align activations from the CNN layers of source and target models.

\subsection{Network Pruning} \label{sec2.2}
Extending the classical work on Optimal Brain Damage~\cite{lecun1990optimal} and Optimal Brain Surgeon~\cite{hassibi1993optimal}, \citeauthor{han2015learning}~(\citeyear{han2015learning}) suggested an iterative training pipeline for model efficiency that alternates between network pruning and fine-tuning.
% there have been several studies that utilize the pruned weights. %\kk{well, network pruning dates back to optimal brain surgeon in 1993} 
In their follow-up work, \cite{han2016dsd} increased the model capacity by removing the sparsity constraint, re-initialized the pruned parameters to zero, and finally re-trained the whole dense network. PackNet \cite{mallya2018packnet}, designed for continual learning (lifelong learning), also employed the pruned parameters to add the new task in the same domain while avoiding catastrophic forgetting \cite{mccloskey1989catastrophic}. In improving the pruned sparse network for higher accuracy, \cite{frankle2018lottery} strongly emphasized the importance of the initial weights determined in the network before pruning. On the other hand, our method is different from the above methods. To reduce the gap between source and target task performances, we suggest the pruned weights should be set to zero and fit the target data with the pruned weights, with regularization to prevent the weight updates deviating too far from zero. 

\section{Proposed Method} \label{sec3}
\paragraph{Formal definition of inductive transfer learning.} Let $\mathcal{X}$ and $\mathcal{Y}$ be the input and output spaces, respectively. We will consider two related but different datasets defined on $\mathcal{X} \times \mathcal{Y}$. The first is the source dataset $\mathcal{D}_S = \{(x_S^i, y_S^i)\}_{i=1}^{N_S}$, where $(x_S^i, y_S^i) \sim p_S(x_S, y_S) = p_S(y_S \vert x_S)p(x_S)$. Similarly, the second is the target dataset ${D}_T = \{(x_T^i, y_T^i)\}_{i=1}^{N_T}$ realized from $p_T(x_T, y_T) = p_T(y_T \vert x_T)p(x_T)$. It should be noted that both datasets share the same input distribution $p(\cdot)$. On the other hand, the conditional distribution $p_S(\cdot \vert x)$ and $p_T(\cdot \vert x)$ are assumed to be different, yet related. The number of target samples is much smaller than that of source samples, i.e., $0 \leq N_{T} \ll N_{S}$. The (supervised) inductive transfer learning is defined as learning $p_{T}(\cdot \vert x)$ (or equivalently the model $f_{T}: \mathcal{X} \to \mathcal{Y}$), by exploiting the knowledge obtained from the source task. For neural network models, it is generally realized by transferring the weights of the source model to the target one, e.g., based on the fine-tuning.
\paragraph{Fine-tuning.} Let $f_{\mathbf{w}_S}: \mathcal{X} \to \mathcal{Y}$ be a pre-trained model on the source dataset $\mathcal{D}_S$ where $\mathbf{w}_S \in \mathbb{R}^D$ denotes $D$-dimensional weight vector of the pre-trained model. Given the target dataset $\mathcal{D}_T$, the fine-tuning method minimizes the standard negative log-likelihood (NLL) $\mathcal{L}_T(\mathbf{w})=-\sum_{i=1}^{N_T} \log p_{\mathbf{w}}(y_T^i \vert x_T^i)$ using the stochastic gradient descent:
\begin{equation} \label{eq1}
\mathbf{w}(t + 1) = \mathbf{w}(t) -\eta \hat{\nabla}_{\mathbf{w}} \mathcal{L}_T(\mathbf{w}), \; \mathbf{w}(0) = \mathbf{w}_S,
\end{equation}
where $\eta$ is a step size and $\hat{\nabla}_{\mathbf{w}} \mathcal{L}_T(\mathbf{w})$ denotes a stochastic estimate of the loss gradient using a mini-batch of data. Thus, the fine-tuning is a maximum likelihood estimation whose the log-prior is centered at $\mathbf{w}_S$.
\paragraph{Starting point (SP) regularization.} Much of the previous work argued that the above simple fine-tuning can lead to a catastrophic forgetting of the source knowledge relevant to the targeted task  \cite{xuhong2018explicit, chen2019catastrophic, li2019delta, li2020baseline, chen2020recall}. It is because some important weights may be driven far away from their initial  $\mathbf{w}_S$. To overcome this issue, \cite{xuhong2018explicit} proposed starting point (SP) regularization that explicitly encourages the transferred target weight to be close to the initial source weight. Specifically, $L^2$-SP regularization is given by:   
\begin{equation*}
\Omega(\mathbf{w}) = \lambda \Vert \mathbf{w} - \mathbf{w}_S \Vert^2 = \lambda \sum_{i=1}^{D} \vert w^i - w_S^i \vert^2,
\end{equation*}
where $\lambda$ is a regularization parameter. With the $L^2$-SP regularizer, the fine-tuning process (\ref{eq1}) is modified as a maximum \textit{a posterior} estimation with the log-prior of $\mathbf{w}_S$:
\begin{equation} \label{eq2}
\mathbf{w}(t + 1) = \mathbf{w}(t) -\eta \hat{\nabla}_{\mathbf{w}} [\mathcal{L}_T(\mathbf{w}) + \Omega(\mathbf{w})], \; \mathbf{w}(0) = \mathbf{w}_S.
\end{equation}
The advantage of SP regularization over the standard fine-tuning is theoretically proved as well as empirically validated \cite{xuhong2018explicit, gouk2020distance, li2021improved}. However, we found that (\ref{eq2}) is still ineffective for transferring the core source knowledge, especially when the amount of the target samples is limited (see experimental results in Section~\ref{sec4}). This is the main motivation behind our work, hypothesizing that keeping the relevant weights \textit{unchanged} is more crucial than constraining all the weights softly. 

\paragraph{All you need is pruning.} Let $\mathbf{w}_U \in \mathbb{R}^K$ and $\mathbf{w}_P \in \mathbb{R}^{D-K}$ be the top-$K$ relevant weights and the remainder, respectively, i.e., $\mathbf{w}(t) = \mathbf{w}_U \oplus \mathbf{w}_P(t)$. We propose a modified transfer learning scenario of (\ref{eq2}) that only updates $\mathbf{w}_P(t)$ while $\mathbf{w}_U$ fixed at SP:
\begin{equation} \label{eq3}
\begin{split}
&\mathbf{w}_P(t + 1) = \mathbf{w}_P(t) -\eta \hat{\nabla}_{\mathbf{w}_P} [\mathcal{L}_T(\mathbf{w}) + \Omega(\mathbf{w}_P)], \\
&\mathbf{w}(0) = \mathbf{w}_U + \mathbf{w}_P(0), \; \Omega(\mathbf{w}_P) = \lambda \sum_{i=1}^{D-K} \vert w_P^i - w_p^i(0) \vert^2.
\end{split}
\end{equation}
Note that the choice of $\mathbf{w}_U$ is highly important for (\ref{eq3}). An ideal situation would be that the weight space is decomposed with respect to the source knowledge, i.e., all of the source information should be embedded in $\mathbf{w}_U$ before the transfer learning. In other words, \textit{$f_{\mathbf{w}_U}$ should be the equivalent sub-network of $f_{\mathbf{w}_S}$}. Meanwhile, the remainder $\mathbf{w}_P$ should have enough capacity to learn the target task. 

Fortunately, the Lottery Ticket Hypothesis (LTH) states that there is an equivalent ($\mathcal{L}_S(\mathbf{w}_U) \leq \mathcal{L}_S(\mathbf{w}_S)$ where $\mathcal{L}_S$ is NLL for the source task) yet efficient ($K\ll D$) sub-network representation of the source model \cite{frankle2018lottery}. In addition, there are a number of notable works that theoretically guarantee LTH or suggest pruning techniques to find such a sub-network practically \cite{malach2020proving, lubana2020gradient, zhang2021validating}. 

\paragraph{PAC-Net.} Inspired by the above, we propose PAC-Net: firstly Prune the source model, secondly Allocate the unpruned ($\mathbf{w}_U$) and pruned weights ($\mathbf{w}_P$) for fixed and learnable parts with respect to the target task, and finally Calibrate\footnote{We use the term \textit{calibration} to mean learning the target task (with few samples), not to be confused with the \textit{calibrated confidence}.}  $\mathbf{w}_P$ via (\ref{eq3}) with few target data. In the following paragraphs, we explain how PAC-Net operates for each step in detail.

\paragraph{Step 1: Pruning.} In this paper, we use the magnitude-based pruning as a baseline method \cite{han2015learning, lubana2020gradient}. It prunes the weight vector $\mathbf{w}_S = (w_S^1, ..., w_S^D)^T$ of the pre-trained model $f_{\mathbf{w}_S}$ by applying the following binary mask $\mathbf{m} = (m^1, ..., m^D)^T$ that keeps the top-$K$ large-magnitude weights:
\begin{equation*}
m^{i} = 
\begin{cases}
1, & \text{if } \vert w^{i} \vert >  w_{\kappa}\\
0, & \text{otherwise,}
\end{cases}
\end{equation*}
where $w_{\kappa}$ is a threshold value determined by $K$. We summarize the pruning step in Algorithm 1.

\begin{algorithm}[h] \label{alg1}
    \SetAlgoLined
    \textbf{Input:} pre-trained weight $\mathbf{w}$ \\
    $w_{\kappa} = \texttt{sort}(\vert \mathbf{w} \vert)_{K}$;\\
    $\mathbf{m} = \mathbbm{1}(\vert \mathbf{w} \vert - w_{\kappa} )$;\\
    \textbf{Output:} pruning mask $\mathbf{m}$;
    \caption{Pruning}
\end{algorithm}

\paragraph{Step 2: Allocation.} Because all the information on the source task should be embedded in the unpruned weights $\mathbf{w}_U = \mathbf{w} \odot \mathbf{m}$, we re-train the masked neural network with fixing:

\begin{equation}
    \begin{split}
\mathbf{w}_U(t + 1) &= \mathbf{w}_U(t) - \eta \hat{\nabla}_{\mathbf{w}_U} \mathcal{L}_S(\mathbf{w}_U), \; \\
\mathbf{w}_U(0) &= \mathbf{w}_S \odot \mathbf{m},
    \end{split}
\end{equation}

After this re-training procedure, one can get the efficient source model $f_{\mathbf{w}_U}$ that will be used as a baseline knowledge on the target task. Namely, we allocate the unpruned part $\mathbf{w}_U$ for the source knowledge; the remained pruned weight $\mathbf{w}_P$ will be used for the target knowledge in the following step. We summarize the allocation step in Algorithm 2.

\begin{algorithm}[h] \label{alg2}
    \SetAlgoLined
    \textbf{Input:} pre-trained weight $\mathbf{w}$, pruning mask $\mathbf{m}$, source dataset $\mathcal{D}_S$, source task loss function $\mathcal{L}_S$, step size $\eta$;\\
    \While{not converged}{
        $\mathbf{w} \leftarrow \mathbf{w} \odot \mathbf{m}$;\\
        $\mathbf{w} \leftarrow \mathbf{w} - \eta \hat{\nabla}_{\mathbf{w}} 
        \mathcal{L}_S(\mathbf{w; \mathcal{D}_S})$;
    }
    $\mathbf{w} \leftarrow \mathbf{w} \odot \mathbf{m}$; \\
    \textbf{Output:} allocated weight $\mathbf{w}$;
    \caption{Allocation}
\end{algorithm}

\paragraph{Step 3: Calibration.} As the last step, PAC-Net adjusts the model with the target data based on the previous pruned model. Our objective is to calibrate the pruned weights $\mathbf{w}_P = \mathbf{w} \odot (\mathbf{1} - \mathbf{m})$ by fitting the model for the target task. Note that the original source weight $\mathbf{w}_U$ should not be updated in the current step to keep the source knowledge completely (hard $L^2$-SP constraint). Meanwhile, $\mathbf{w}_P$ is updated as follows (soft $L^2$-SP constraint): 
\begin{equation} \label{eq4}
\begin{split}
&\mathbf{w}_P(t + 1) = \mathbf{w}_P(t) - \eta \hat{\nabla}_{\mathbf{w}_P} [\mathcal{L}_T(\mathbf{w}) + \Omega(\mathbf{w}_P)], \\
&\mathbf{w}(0) = \mathbf{w}_U, \; \Omega(\mathbf{w}_P) = \lambda \Vert \mathbf{w}_P \Vert = \lambda \sum_{i=1}^{D-K} \vert w_P^i \vert^2.
\end{split}
\end{equation}
 We summarize the calibration step in Algorithm 3. It is noteworthy that (\ref{eq4}) is equal to (\ref{eq3}) based on the fact that the SP of $\mathbf{w}_P$, i.e., $\mathbf{w}_P(0) = (w_P^1(0), ..., w_P^{D-K}(0))^T$, should be $\mathbf{0}$ considering the baseline source model is given by $f_{\mathbf{w} \odot \mathbf{m}} = f_{\mathbf{w}_U \oplus \mathbf{0}}$. We use this $L^2$ regularization with zero SP of $\mathbf{w}_P$ as a baseline; we empirically found that this baseline PAC-Net outperforms other competitors including the fine-tuning, (soft) $L^2$-SP and (semi-hard) $L^2$-SP-Fisher ($L^2$-SP with Fisher information), and (hard) PAC-Nets initialized with different SP of $\mathbf{w}_P$. Lastly but not least, PackNet \cite{mallya2018packnet}, designed for continual learning, used the same strategy with ours, except for the $L^2$ regularization. Note that this is a special case of PAC-Net when the target task is too remotely relevant for inductive transfer learning despite the importance of the regularization demonstrated by \cite{xuhong2018explicit}.
 
\begin{algorithm}[h] \label{alg3}
    \SetAlgoLined
    \textbf{Input:} neural network $f$, allocated weight $\mathbf{w}$, pruning mask $\mathbf{m}$, target dataset $\mathcal{D}_T$, target task loss function $\mathcal{L}_T$, SP regularizer $\Omega$, step size $\eta$;\\
    $\mathbf{w}_U =  \mathbf{w} \odot \mathbf{m}$ \\
    \While{not converged}{
        $\mathbf{w} \leftarrow \mathbf{w} \odot (\mathbf{1} - \mathbf{m}) + \mathbf{w}_U$;\\
        $\mathbf{w} \leftarrow \mathbf{w} - \eta \hat{\nabla}_{\mathbf{w}} [\mathcal{L}_T(\mathbf{w}) + \Omega(\mathbf{w} \odot (\mathbf{1} - \mathbf{m}))]$;
    }
    $\mathbf{w}_P = \mathbf{w} \odot (\mathbf{1} - \mathbf{m})$; \\
    \textbf{Output:} target model $f_{\mathbf{w}_T} = f_{\mathbf{w}_U + \mathbf{w}_P}$
    \caption{Calibration}
\end{algorithm}

\section{Experiments} \label{sec4}
In this section, we evaluate the proposed method with various problems on inductive transfer learning. Section \ref{sec4.1} compares PAC-Net with various inductive transfer learning algorithms on the regression problem, which can control a distance between the source and the target task. Section \ref{sec4.2} compares PAC-Net to the multiple algorithms on the classification dataset. Section \ref{sec4.3} - \ref{sec4.5} assess PAC-Net on the real-world scenarios, which have to close the gap between simulation and reality. Section \ref{sec4.3} and Section \ref{sec4.4} evaluate inductive transfer learning algorithms with a neural network that trains ordinary differential equations (ODEs) or partial differential equations (PDEs). Section \ref{sec4.5} applies PAC-Net to the real-world problem, which consists of complex multi-physics. Note that information on the whole experiments in this section is described in Appendix \ref{app:1}.

\begin{table*}[t!] 
\begin{center}
\caption{\textbf{Comparison of different approaches to solve a modified Friedman \#1 problem for inductive transfer learning.} Each score is an averaged Root Mean-Squared Errors (RMSEs) over five runs with varying the target samples from 10 to 100. $d$ is a distance between the source and target task.} \label{table:1}
\vskip 0.1in
{
\footnotesize
\begin{tabular}{cccccccccccc}

\toprule

$d$ & Method & 10 & 20 & 30 & 40 & 50 & 60 & 70 & 80 & 90 & 100\\
\midrule
\multirow{6}{*}{1} & Target only & 19.2 & 12.8 & 12.9 & 10.9 & 10.6 & 10.0 & 9.8 & 9.9 & 9.1 & 9.4 \\
& Boosting & 33.8 & 26.3 & 28.4 & 27.5 & 24.1 & 29.6 & 24.6 & 26.7 & 28.4 & 28.1 \\
 & Fine-tuning & 10.8 & 8.8 & 7.7 & 7.2 & 6.6 & 6.3 & 6.0 & 5.8 & 5.6 & 5.4 \\
 & $L^2$-SP & 12.5 & 10.1 & 8.8 & 7.9 & 7.3 & 7.0 & 6.5 & 6.3 & 6.0 & 5.8 \\
  & $L^2$-SP-Fisher & 13.5 & 11.1 & 10.0 & 9.2 & 9.2 & 8.7 & 8.4	& 8.4 & 8.0 & 7.9 
 \\
  & PAC-Net & \textbf{6.7} & \textbf{5.8} & \textbf{5.1} & \textbf{4.6} & \textbf{4.1} & \textbf{4.0} & \textbf{3.7} & \textbf{3.6} & \textbf{3.4} & \textbf{3.3}
 \\
\midrule
\multirow{6}{*}{2} & Target only & 89.0 & 49.0 & 43.8 & 38.0 & 34.9 & 33.6 & 33.2 & 33.9 & 33.1 & 31.7  \\
& Boosting & 146.9 & 151.1 & 138.8 & 135.5 & 143.1 & 134.6 & 139.1 & 138.6 & 135.5 & 144.5  \\
 & Fine-tuning & 75.9 & 47.9 & 39.5 & 32.9 & 31.0 & 31.0 & 30.3 & 29.3 & 28.6 & 27.1 \\
 & $L^2$-SP & 69.2 & 41.2 & 29.8 & 26.3 & 25.8 & 25.7 & 24.5 & 23.3 & 23.6 & 22.0 
 \\
  & $L^2$-SP-Fisher & 71.5 & 39.3 & 29.1 & 25.8 & 25.7 & 25.7 & 24.8 & 24.1 & 24.2 & 24.6
  \\
  & PAC-Net & \textbf{52.4} & \textbf{31.0} & \textbf{25.1} & \textbf{23.4} & \textbf{22.1} & \textbf{21.1} & \textbf{21.0} & \textbf{20.8} & \textbf{20.1} & \textbf{19.0}\\
\midrule
\multirow{6}{*}{3} & Target only & 270.8 & 143.3 & 112.0 & 89.3 & 85.8 & 81.5 & 76.4 & 69.7 & 70.9 & 64.0 \\
& Boosting & 509.7 & 495.9 & 515.4 & 495.8 & 489.8 & 475.7 & 486.2 & 468.2 & 472.8 & 457.4 \\
 & Fine-tuning & 227.4 & 125.1 & 98.0 & 79.9 & 74.6 & 72.9 & 71.8 & 66.0 & 63.8 & 59.0 \\
 & $L^2$-SP & 217.4 & 92.3 & 71.4 & 54.8 & 52.3 & 48.3 & 49.1 & 44.7 & 43.5 & 39.9 \\
 & $L^2$-SP-Fisher & 224.9 & \textbf{86.4} & 68.1 & 52.8 & 53.6 & 52.5 & 48.9 & 46.1 & 43.4 & 43.6

 \\
  & PAC-Net & \textbf{186.2} & 88.9 & \textbf{62.7} & \textbf{52.2} & \textbf{47.2} & \textbf{46.1} & \textbf{42.7} & \textbf{42.3} & \textbf{40.0} & \textbf{37.4}\\

\bottomrule
\end{tabular} 
\vskip -0.4in
}
\end{center}

\end{table*}

\begin{table*}[t!] 
\begin{center}
\vskip -0.2in
\caption{\textbf{The results of ablation studies on the modified Friedman \#1 problem for inductive transfer learning.} Each score is an averaged RMSEs over five runs with varying the target samples from 10 to 100.} \label{table:2}
\vskip 0.1in
{
\footnotesize
\begin{tabular}{ccccccccccccc}

\toprule

$d$ & Method & 10 & 20 & 30 & 40 & 50 & 60 & 70 & 80 & 90 & 100 & Avg.\\
\midrule
\multirow{6}{*}{1} & PC-Net & 11.4 & 9.1 & 8.4 & 7.6 & 7.2 & 6.8 & 6.3 & 6.2 & 5.9 & 5.7 & 7.5
 \\
 & PAC-Net (No-$L^2$) & 9.6 & 7.5 & 6.1 & 5.7 & 5.3 & 5.1 & 4.6 & 4.4 & 4.3 & 4.1 & 5.7 
 \\
 & PAC-Net (RI) & 7.1 & 7.2 & 6.1 & 5.4 & 5.4 & 5.5 & 4.9 & 4.4 & 5.1 & 4.6 & 5.6

 \\
 & PA-Net-$L^2$-SP & 10.6 & 8.9 & 7.6 & 6.6 & 5.8 & 5.2 & 4.7 & 4.6 & 4.4 & 4.2 & 6.3

 \\
 & PA-Net-$L^2$-SP-Fisher & 12.3 & 10.4 & 9.7 & 9.2 & 8.6 & 8.7 & 8.4 & 8.0 & 8.0 & 7.8 & 9.1 
 \\
  & PAC-Net & \textbf{6.7} & \textbf{5.8} & \textbf{5.1} & \textbf{4.6} & \textbf{4.1} & \textbf{4.0} & \textbf{3.7} & \textbf{3.6} & \textbf{3.4} & \textbf{3.3} & \textbf{4.4}
 \\
\midrule
\multirow{6}{*}{2} & PC-Net & 61.8 & 33.6 & 27.9 & 24.2 & 23.1 & 22.5 & 22.5 & 22.3 & 22.1 & 20.8 & 28.1

\\
& PAC-Net (No-$L^2$) & 61.6 & 38.4 & 30.4 & 27.6 & 25.7 & 26.2 & 25.5 & 25.4 & 25.6 & 23.8 & 31.0 

 \\
 & PAC-Net (RI) & \textbf{50.7} & 31.1 & \textbf{25.1} & 23.9 & 22.1 & 21.3 & 21.2 & 21.0 & 20.6 & 19.9 & 25.7

 \\
 & PA-Net-$L^2$-SP & 65.1 & 31.9 & 25.9 & \textbf{22.7} & \textbf{21.4} & \textbf{20.9} & 21.3 & \textbf{20.4} & 20.2 & 20.1 & 27.0

 \\
 & PA-Net-$L^2$-SP-Fisher & 70.8 & 37.8 & 28.0 & 24.9 & 26.5 & 24.9 & 24.5 & 26.7 & 23.8 & 23.9 & 31.2 

 \\
  & PAC-Net & 52.4 & \textbf{31.0} & \textbf{25.1} & 23.4 & 22.1 & 21.1 & \textbf{21.0} & 20.8 & \textbf{20.1} & \textbf{19.0} & \textbf{25.6}\\
\midrule
\multirow{6}{*}{3}  & PC-Net & 212.1 & 106.3 & 81.6 & 63.1 & 59.3 & 55.9 & 56.0 & 52.5 & 52.3 & 47.5 & 78.7

 \\
 & PAC-Net (No-$L^2$) & 208.0 & 115.6 & 88.0 & 72.0 & 68.7 & 64.9 & 59.9 & 56.5 & 56.7 & 53.3 & 84.4

 \\
 & PAC-Net (RI) & 192.5 & 99.8 & 67.0 & 58.4 & 54.8 & 54.2 & 52.3 & 48.3 & 47.0 & 45.0 & 71.9

 \\
 & PA-Net-$L^2$-SP & 207.4 & 89.8 & 61.4 & \textbf{47.7} & \textbf{45.2} & \textbf{44.9} & 44.6 & \textbf{40.9} & \textbf{38.6} & 38.3 & 65.9 

 \\
 & PA-Net-$L^2$-SP-Fisher & 230.9 & \textbf{88.0} & \textbf{54.8} & 53.2 & 46.5 & 47.5 & 48.3 & 45.1 & 40.3 & 43.2 & 69.8 

 \\
  & PAC-Net & \textbf{186.2} & 88.9 & 62.7 & 52.2 & 47.2 & 46.1 & \textbf{42.7} & 42.3 & 40.0 & \textbf{37.4} & \textbf{64.6}\\

\bottomrule
\end{tabular} 
\vskip -0.4in
}
\end{center}
\end{table*}

\subsection{Regression} \label{sec4.1}
Friedman \#1 (Friedman, 1991) is a well-known regression problem and \cite{boostingforreg} modified the problem for inductive transfer learning. We customized the problem to be similar to \cite{boostingforreg} but to be more distinguishable between source and target tasks. Each instance $\mathbf{x} = (x_1, ..., x_{10})$ is a feature vector of length ten, where each component $x_i$ is drawn independently from the uniform distribution [0, 1]. The label for each instance is dependent only on the first five features:
\begin{equation*}
\begin{split}
y =
& \, a_1 \cdot 10\sin(\pi(b_1x_1+c_1) \cdot (b_2x_2+c_2)) +\\
& \, a_2 \cdot 20(b_3x_3+c_3-0.5)^2 + a_3 \cdot 10(b_4x_4+c_4) +\\
& \, a_4 \cdot 5(b_5x_5+c_5) + \mathcal{N}(0,s),
\end{split}
\end{equation*}
where $\mathcal{N}$ is the normal distribution and each $a_i$, $b_i$ and $c_i$ represents a fixed parameter. We used $a_i$, $b_i$ are set to 1 and $c_i$ is set to 0 for source while $a_i$, $b_i$, $c_i$ are set to $d$ for the target. Therefore, the larger $d$ is, the further the distance of source and target becomes. Furthermore, we injected noise terms for target tasks to reflect the real-world problem.

We compared our method with Boosting \cite{boostingforreg}, Fine-tuning \cite{FT}, $L^2$-SP, and $L^2$-SP-Fisher \cite{xuhong2018explicit}.
Table \ref{table:1} shows that our method produces the best performance in almost experiment. Compared to $L^2$-SP and $L^2$-SP-Fisher, Table \ref{table:1} experimentally demonstrates keeping the source weights as a hard constraint helps to train the target task. Boosting shows a worse performance than a baseline algorithm, which only trains the target dataset (Target only) because boosting method over-fits residual errors.

\begin{table*}[t!] 
\begin{center}
\caption{\textbf{Comparison of different approaches for a binary classification task on CelebA dataset.} For almost cases with varying the number of target samples, PAC-Net achieves higher accuracy than other methods. All scores are the average accuracy of five runs.} \label{table:3}
\vskip 0.1in
{
\small
\begin{tabular}{ccccccccccccc}
\toprule

S to T & Method & 10 & 20 & 30 & 40 & 50 & 60 & 70 & 80 & 90 & 100 & Avg.\\

\midrule
\multirow{6}{*}{\shortstack{Arched\\Eyebrows\\$\downarrow$\\Eyeglasses}} & Target only & 60.7 & 56.5 & 60.4 & 64.8 & 63.9 & 66.1 & 66.3 & 66.8 & 66.9 & 66.0 & 63.8
 \\
 & Fine-tuning & 54.9 & 49.8 & 57.5 & 64.5 & 67.6 & 67.3 & 67.5 & 66.8 & 69.0 & 71.8 & 63.7
 \\
 & $L^2$-SP & 57.9 & 55.2 & 56.8 & 59.2 & 61.0 & 59.7 & 67.3 & 61.8 & 61.6 & 67.5 & 60.8
 \\
  & $L^2$-SP-Fisher & \textbf{65.2} & 62.0 & 61.7 & 67.6 & 68.2 & 67.1 & 68.9 & 69.1 & 69.9 & 69.3 & 66.9
  \\
  & DELTA & 55.2 & 52.8 & 56.1 & 66.1 & 67.5 & 67.1 & \textbf{71.2} & 71.3 & 68.1 & \textbf{75.7} & 65.1
 
 \\
  & PAC-Net & 64.1 & \textbf{63.2} & \textbf{66.6} & \textbf{70.2} & \textbf{72.8} & \textbf{70.5} & 70.7 & \textbf{73.5} & \textbf{71.2} & 72.6 & \textbf{69.5}

 \\
\midrule

\multirow{6}{*}{\shortstack{Eyeglasses\\$\downarrow$\\Arched\\Eyebrows}} & Target only & 57.7 & 63.4 & 64.9 & 60.8 & 67.8 & 69.1 & 61.6 & 68.6 & 70.9 & 66.7 & 65.1 
  \\
& Fine-tuning & 52.4 & 61.7 & \textbf{67.7} & 70.8 & 72.0 & 74.0 & 74.2 & 74.2 & 74.7 & 75.3 & 69.7
 \\
 & $L^2$-SP & 60.6 & 57.8 & 58 & 66.5 & 62.8 & 60.3 & 73.7 & 64.6 & 63.0 & \textbf{81.4} & 64.9
 \\
  & $L^2$-SP-Fisher & 62.9 & 62.7 & 64.7 & 70.6 & 70.6 & 68.6 & 73.0 & 72.6 & 72.7 & 73.2 & 69.1
 \\
  & DELTA & 59.3 & 58.0 & 57.2 & 64.0 & 63.1 & 62.3 & 63.3 & 64.4 & 64.4 & 63.9 & 62.0
 \\
  & PAC-Net & \textbf{70.0} & \textbf{64.9} & 65.8 & \textbf{79.5} & \textbf{78.8} & \textbf{77.5} & \textbf{78.6} & \textbf{80.1} & \textbf{78.8} & 80.4 & \textbf{75.5}
\\
\bottomrule

\end{tabular} 
}
\end{center}

\end{table*}

\begin{figure*}[t!]
\begin{center}
\end{center}
   \centering
      \vskip -0.15in
   \includegraphics[width=0.9\linewidth]{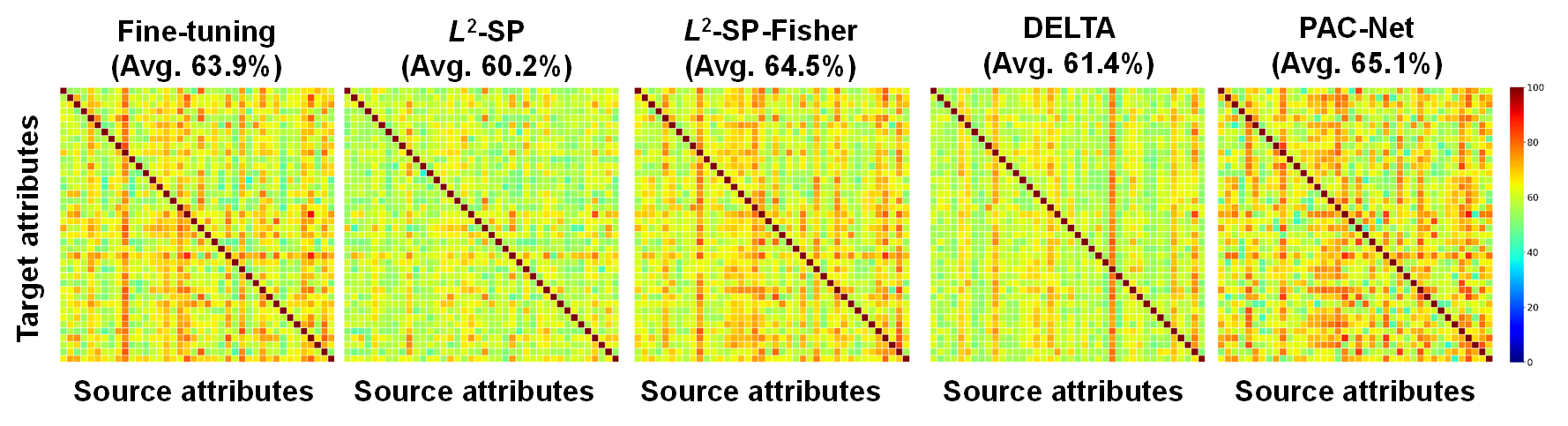}

   \caption{
   \textbf{Accuracy of the multiple algorithms with 50 target samples between individual attributes on the CelebA dataset.} PAC-Net clearly shows the best performance in varying the attributes between the source and the target. Each attribute is in alphabetical order and is not indicated for readability.
   } \label{fig:3}
\end{figure*}

We further conducted an ablation study to appreciate each steps of pruning, allocation, and calibration in PAC-Net. Isolating each step yields the following algorithms:

\begin{itemize}
    \item PC-Net: We fix $\mathbf{w}_U$ collected by pruning and we calibrate $\mathbf{w}_P$ to the target task while skipping the allocation step, i.e., $\mathbf{w}_U$ is NOT updated.
    \item PAC-Net (No-$L^2$): We calibrate $\mathbf{w}_P$ to the target task without $L^2$ regularization. This is essentially PackNet \cite{mallya2018packnet}.
    \item PAC-Net (RI): We initialize $\mathbf{w}_P$ to random values in the calibration step.
    \item PA-Net-$L^2$-SP or PA-Net-$L^2$-SP-Fisher: After the pruning and allocation steps, we do not fix $\mathbf{w}_U$ but calibrate all weights to the target task with soft ($L^2$-SP) or semi-hard regularization ($L^2$-SP-Fisher).
\end{itemize}

Table \ref{table:2} summarizes the whole ablation studies. Comparing PAC-Net to PC-Net, we confirmed the importance of keeping the complete source knowledge. This result is consistent with $L^2$-SP and $L^2$-SP-Fisher that partially forget the source knowledge. As a result compared to PAC-Net (No-$L^2$), we proved it is crucial to regularize $\mathbf{w}_P$ although preserving the source knowledge. Consistent with the hypothesis of $L^2$-SP, zero SP of $\mathbf{w}_P$ shows the better performance than the random SP of $\mathbf{w}_P$ (PAC-Net (RI)).

\subsection{Classification} \label{sec4.2}
CelebFaces Attributes Dataset (CelebA)~\cite{celeba} is a large-scale celebrity images with the forty attribute annotations. For the source task, we trained the ResNet-18 \cite{resnet} with one attribute as a binary classification, and then fitted the trained model to the target tasks with different attributes by varying the number of target samples. Table \ref{table:3} is an example describing whether the model that trains an eyeglasses attribute can help to train the arched eyebrow attribute, or vice versa. The baseline algorithms for comparison are fine-tuning, $L^2$-SP, $L^2$-SP-Fisher, and DELTA \cite{li2019delta}.
Our method shows the better performance than others. For all the forty attributes, we examined the performance of different approaches for the binary classification tasks. Figure \ref{fig:3} clearly indicates our method outperforms other algorithms.

\begin{figure*}[t!]

\begin{center}
\end{center}
   \centering
   \includegraphics[width=0.97\linewidth]{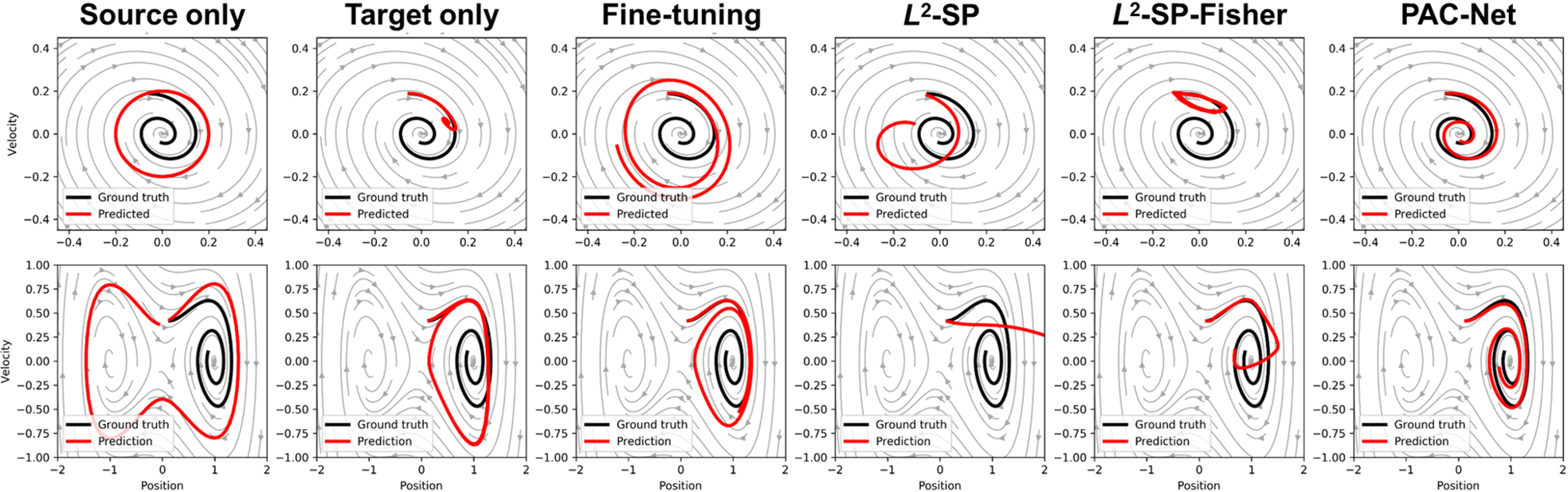}

   \caption{
   \textbf{Phase plot of the multiple algorithms with Neural ODE for linear oscillator (top) and non-linear oscillator (bottom).} The linear and non-linear oscillator are the results with three and eight target samples, respectively. PAC-Net can calibrate the discrepancy between the source and the target while others cannot.} \label{fig:4}

\end{figure*}

\begin{table*}[t!] 
\begin{center}

\caption{\textbf{RMSE results of multiple inductive transfer learning algorithms with Neural ODE model on Duffing equation.} PAC-Net shows the lowest error in most experiments.} \label{table:4}
\vskip 0.1in
{
\small
\begin{tabular}{ccccccccccccc}
\toprule

System & Method & 10 & 20 & 30 & 40 & 50 & 60 & 70 & 80 & 90 & 100 & Avg.\\
\midrule
\multirow{5}{*}{Linear} & Target only & 2.85  & 3.30  & 0.54  & 0.42  & 0.36  & 0.84  & 0.16  & 0.10  & 0.09  & 0.06  & 0.87 

 \\
 & Fine-tuning & 3.88  & 0.53  & 0.60  & 0.18  & 0.12  & \textbf{0.05} & 0.05  & \textbf{0.03} & \textbf{0.02} & \textbf{0.02} & 0.55 

 \\
 & $L^2$-SP & 0.84  & 0.77  & 0.68  & 0.65  & 0.58  & 0.58  & 0.57  & 0.57  & 0.56  & 0.52  & 0.63 
\\
 & $L^2$-SP-Fisher & 3.43 & 0.82 & 0.23 & 0.18 & 0.07 & \textbf{0.05} & 0.06 & 0.05 & 0.07 & 0.04 & 0.5

 \\
  & PAC-Net & \textbf{0.19}  & \textbf{0.11} & \textbf{0.06} & \textbf{0.05} & \textbf{0.05} & \textbf{0.05} & \textbf{0.04} & 0.04 & 0.03 & 0.03 & \textbf{0.07}

 \\
\midrule

\multirow{5}{*}{Non-linear} & Target only & 3.23  & 3.40  & 1.49  & 0.85  & 1.18  & 1.05  & 1.07  & 0.51  & 0.44  & 0.33  & 1.36

  \\
 & Fine-tuning & \textbf{0.73}  & 2.16  & \textbf{0.62} & 0.85  & 0.31  & 0.41  & 0.26  & 0.24  & 0.25  & \textbf{0.19} & 0.60

 \\
  & $L^2$-SP & 0.84  & 1.59  & 1.70  & 1.91  & 1.86  & 2.21  & 2.24  & 2.38  & 1.93  & 2.12  & 1.88 
 \\
& $L^2$-SP-Fisher & 3.46 & 1.56 & 1.10 & 1.04 & 1.39 & 1.52 & 1.15 & 1.15 & 0.49 & 0.4 & 1.33

 \\
  & PAC-Net & 0.92  & \textbf{0.76} & 0.64 & \textbf{0.65} & \textbf{0.22} & \textbf{0.21} & \textbf{0.20} & \textbf{0.21} & \textbf{0.20} & 0.20 & \textbf{0.42} \\

\bottomrule

\end{tabular} 

}
\end{center}
\end{table*}

\subsection{Ordinary Differential Equations} \label{sec4.3}
We introduce another task obeying ordinary differential equations as the real-world scenario. We introduce Duffing oscillator \cite{kovacic2011duffing} given by:
\begin{equation*}
    \frac{d\mathbf{q}}{dt} = \mathbf{p}, \frac{d\mathbf{p}}{dt} = - \alpha\mathbf{q} - \beta\mathbf{q}^3 - \gamma\mathbf{p}
\end{equation*}
where $\alpha, \beta$, and $\gamma$ are the calibration parameters that control the linear stiffness, the amount of non-linearity in the restring force, and the amount of damping, respectively. $\mathbf{p}$ and $\mathbf{q}$ are position and velocity, respectively. There are two system, linear and non-linear oscillator for inductive transfer learning problems.

\paragraph{Linear oscillator.} We choose a linear oscillator, i.e., $\alpha = 1, \beta = 0, \gamma = 0$ in the source task and $\alpha = 1, \beta = 0, \gamma = 0.3$ in the target task, respectively.
\paragraph{Non-linear oscillator.} As the more complex problem, we introduced the non-linear oscillator, i.e., $\alpha = -1, \beta = 1, \gamma = 0$ in the source task and $\alpha = -1, \beta = 1, \gamma = 0.3$ in the target task, respectively.

With two oscillation systems, the source task denotes the energy conservation system and the target task the energy loss system caused by the frictional force, as described in Section \ref{sec1}. In this task, we should infer the trajectory in the reality (target task) after the observed trajectory within a few seconds. The leftmost of Figure \ref{fig:4} shows the discrepancy between the simulation (source) and experiment (target). Note that as a base learner, we employed a Neural ODE \cite{nodes} to predict the trajectory. Figure \ref{fig:4} depicts PAC-Net can well calibrate the target task based on the source task while others cannot. Table \ref{table:4} describes RMSE of the target trajectory. PAC-Net predicts the most accurate trajectory in both systems over most experiments that varies the number of target data.

\subsection{Partial Differential Equations} \label{sec4.4}
We evaluate PAC-Net with real-world scenarios in which the discrepancy between the simulation and reality should be close. The diffusion equation is one of PDEs with various applications. We choose the equation with Dirichlet boundary condition, as follows:

\begin{equation} \label{eq6}
    \begin{split}
        & \partial_t u(x, y, t) = v \Delta u(x,y,t), x,y \in (0,2), t \in (0, 1] \\
        & \text{subject to initial condition:} \\
        & u(x, y, 0) = 
        \begin{cases}
            2 & \text{if 0.5 $<$ x, y $<$ 1} \\
            1 & \text{Otherwise}
        \end{cases} 
        \\
        & \text{and boundary condition:}
        \\
        & u(0, y, t) = u(2, y, t) = u(x, 0, t) =  u(x, 2, t) =  1,  
    \end{split}
\end{equation}

\begin{figure}[t!]
\begin{center}
\end{center}
   \centering
   \includegraphics[width=0.9\linewidth]{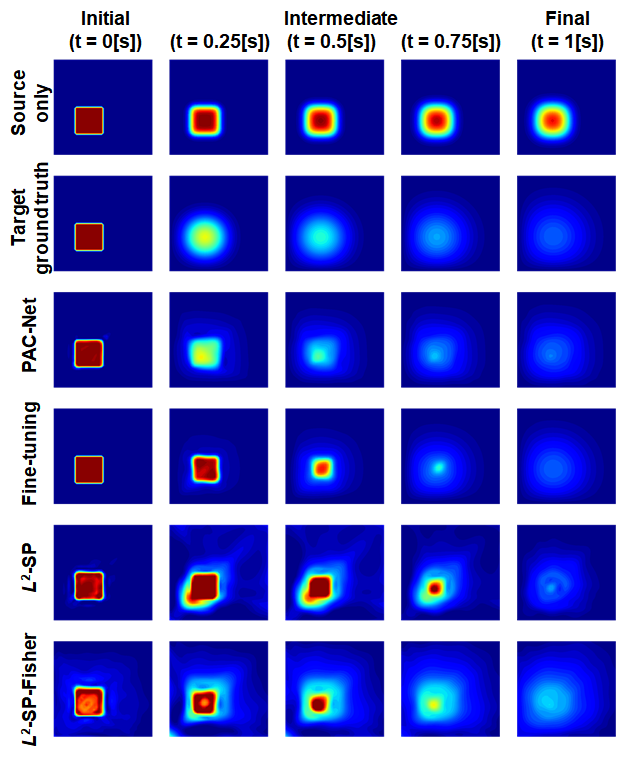}
   \caption{
   \textbf{Diffusion flows of various algorithms with PINN as time advances.} PAC-Net accurately predicts the concentration flows at a intermediate time, which has never been so far.
   } \label{fig:5}
\end{figure}

\begin{table}[t] 
\begin{center}
\caption{\textbf{RMS error on diffusion equation in the target task.} } \label{table:5}
{
\small
\begin{tabular}{cccccc}
\toprule
Method & 0.25 [s] & 0.5 [s] & 0.75 [s] & 1.0 [s] & Avg.\\
\midrule
Fine-tuning & 0.164 & 0.104 & 0.028 & \textbf{0.001} & 0.074\\
$L^2$-SP & 0.396 & 0.310 & 0.132 & 0.023 & 0.215\\
$L^2$-SP-Fisher & 0.192 & 0.162 & 0.108 & 0.089 & 0.138\\
PAC-Net & \textbf{0.056} & \textbf{0.039} & \textbf{0.020} & 0.006 & \textbf{0.030}\\
\bottomrule

\end{tabular} 

}
\end{center}
\vskip -0.2in
\end{table}

where $u$ denotes the concentration, $v$ is an unknown parameter that can vary with the materials. We set $v$ to be 0.01 and 0.1 for the source and target task, respectively. With only initial and the boundary conditions (\ref{eq6}) in the source task, we trained a physics-informed neural network (PINN)~\cite{pinn}, which learns PDEs of (\ref{eq6}). As a realistic scenario\cite{chh}, we assume that final concentration data with the unknown $v$ is only available so that we make model predict a concentration profile at intermediate steps. As shown in Figure \ref{fig:5}, the concentration for target task shows a considerable discrepancy with that for source task. We fitted PINN that learned the diffusion equation in the source task to the initial and final concentration in the target task with various inductive transfer learning algorithms. Figure \ref{fig:5} clearly shows that PAC-Net can predict the target data at intermediate step while others do not. Table \ref{table:5} shows the mean-squared error as time advances, which depicts PAC-Net outperforms others.

\begin{figure}[t!]
\begin{center}
\end{center}
   \centering
   \includegraphics[width=0.9\linewidth]{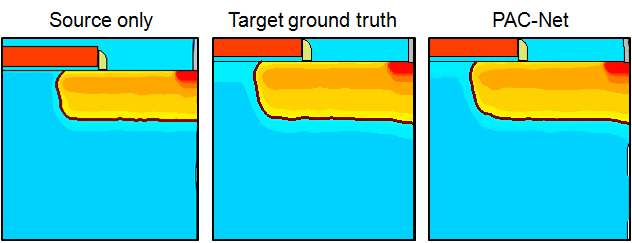}

   \caption{
   \textbf{Prediction results with 40 target samples.}} \label{fig:6}
\vskip -0.2in
\end{figure}

\subsection{Real-world Problem} \label{sec4.5}
In this section, we introduce semiconductor dataset, as a real-world problem that experiences a difference between reality and simulation. The dataset consists of the scalar values as an input and images as an output. We used the RTT model (\citeauthor{rtt},~\citeyear{rtt},~\citeyear{rtt2}) to solve this problem. The left and the center of Figure \ref{fig:6} illustrate the outputs of source and target tasks can differ although the input features are the same. Table \ref{table:6} shows PAC-Net works well even when only 40 target samples are available in training. Figure \ref{fig:6} indicates that PAC-Net can clearly predict the target task.

\begin{table}[t!] 
\begin{center}
\caption{\textbf{Results with RTT model on semiconductor dataset.}} \label{table:6}
\vskip 0.1in
{
%\scriptsize
\small
\begin{tabular}{cccccc}
\toprule
Method & 20 & 40 & 60 & 80 & 100 \\
\midrule
Target only & 0.703 & 0.799 & 0.855 & 0.883 & 0.893 \\
PAC-Net & \textbf{0.884} & \textbf{0.907} & \textbf{0.924} & \textbf{0.932} & \textbf{0.933} \\
\bottomrule
\end{tabular} 

}
\end{center}
\vskip -0.2in
\end{table}

\section{Conclusion} \label{sec5}
Inductive transfer learning aims to resolve the distributional shift problem with a little target data while there exists enough amount of source data. Consistent with previous studies, we found the importance of preserving source knowledge. $L^2$-SP, DELTA, and PAC-Net aim to preserve the knowledge in source tasks while solving the aforementioned problem by taking different approaches. $L^2$-SP imposes the regularization to all weights to preserve the source knowledge, and DELTA additionally applies channel-wise attention to feature maps. On the other hand, PAC-Net encourages updating the insignificant weights while keeping the source weights. Experimental results show that PAC-Net is more effective than competitive baseline methods.

\newpage
\bibliography{main}
\bibliographystyle{icml2022}

%%%%%%%%%%%%%%%%%%%%%%%%%%%%%%%%%%%%%%%%%%%%%%%%%%%%%%%%%%%%%%%%%%%%%%%%%%%%%%%
%%%%%%%%%%%%%%%%%%%%%%%%%%%%%%%%%%%%%%%%%%%%%%%%%%%%%%%%%%%%%%%%%%%%%%%%%%%%%%%
% APPENDIX
%%%%%%%%%%%%%%%%%%%%%%%%%%%%%%%%%%%%%%%%%%%%%%%%%%%%%%%%%%%%%%%%%%%%%%%%%%%%%%%
%%%%%%%%%%%%%%%%%%%%%%%%%%%%%%%%%%%%%%%%%%%%%%%%%%%%%%%%%%%%%%%%%%%%%%%%%%%%%%%
\newpage
\appendix
\onecolumn

\section{Experiment Details} \label{app:1}
\subsection{Regression}
\paragraph{Architectural Details.} We used two fully-connected layers with output size of 200 with ReLU activation, and one fully-connected layer as the prediction function with linear activation. All layers are initialized with he normal \cite{he2015delving}, and the mean-squared error is used as loss function with a batch size of 128 where Adam optimizer \cite{kingma2015adam} is performed with the step size $10^{-4}$. For TradaBoost.R2 \cite{boostingforreg}, which is available on the website\footnote{https://github.com/jay15summer/Two-stage-TrAdaboost.R2.}, we choose the two-stage version with 10 first stage and five second stage iterations. For $L^2$-SP, we choose the regularization parameter to 0.01. For PAC-Net, we set pruning ratio to 0.8 and $\lambda$ to 0.01.
\paragraph{Experiment Details.} We carried out five experiments as follows: i) there are 20,000 datasets, half of which are training sets, and the rest are test sets. ii) input features of both domains are same. iii) target training datasets vary in size. Note that the source model is fixed in every iteration adding target samples.

\subsection{CelebA}
\paragraph{Architectural Details.} We used ResNet-18 architecture. For $L^2$-SP and PAC-Net, we regularized the weights of all convolution layers with the penalty of each method. We imposed no regularization to the dense layer for PAC-Net while applying $L^2$ regularization whose strength is 0.01 to the dense layer for $L^2$-SP. For fine-tuning, we fitted all weights based on the source weights. The parameters we used are as follows: the loss function is cross-entropy, the optimizer Adam with the step size $10^{-4}$, batch size 128. For PAC-Net, we set pruning ratio to 0.8 and $\lambda$ to 0.01. For $L^2$-SP, we set the $L^2$ strength of convolution layers to 0.01.

\paragraph{Experiment Details.} We resized the image to 64x64. For an experiment, we selected each 2000 images for source and target. Also, we randomly picked another 2000 images from the rest of attributes (38 attributes) for model to learn generic knowledge regrading datasets. We used a half of the images for the train dataset and the rest for the test dataset. We repeated the experiments five times with the different random seeds.

\subsection{Ordinary Differential Equations}
\paragraph{Architectural Details.} We used six fully-connected layers with output size of 256 with tanh activation, and one fully-connected layer as the prediction function with linear activation. For Neural ODE \cite{nodes} solver, we used adaptive Runge-Kutta (RK) methods \cite{dopri}. The parameters we used are as follows: the loss function is mean-squared error, the optimizer Adam with the step size $10^{-4}$, batch size 512. For PAC-Net, we set pruning ratio to 0.8 and $\lambda$ to 0.001. For $L^2$-SP, we set the regularization parameter to 0.001.

\paragraph{Experiment Details.} This experiment is similar with \cite{huh2020time}. From equation 1, we generated 100 and 10 trajectories whose time interval is 0.1 for source and target dataset, respectively. For each trajectory, the initial state $(\mathbf{q}(t_0), \mathbf{p}(t_0))$ is uniformly sampled from annulus in [0.2, 1]. For the target task, we added uniformly distributed noise multiplied by 0.01 for each trajectory. With ten trajectories for target task, we assessed the RMSE losses of the trajectories from one second to ten second.

\subsection{Partial Differential Equations}
\paragraph{Architectural Details.} We used six fully-connected layers with output size of 256 with swish activation \cite{swish}, and one fully-connected layer as the prediction function with linear activation. When training the source task, we used the particular loss function that (PINN, \cite{pinn}) proposed, as follows:

For the target task, we used mean-squared-error to fit the model with an available target data. The parameters we used are as follows: the optimizer is Adam with the step size $10^{-4}$, batch size 32. For PAC-Net, we set pruning ratio to 0.8 and $\lambda$ to 0.01. For $L^2$-SP, we set the $L^2$ strength to 0.01.

\subsection{Real-world Problem}
\begin{figure}[h]
\begin{center}
\end{center}
   \centering
   \includegraphics[width=0.5\textwidth]{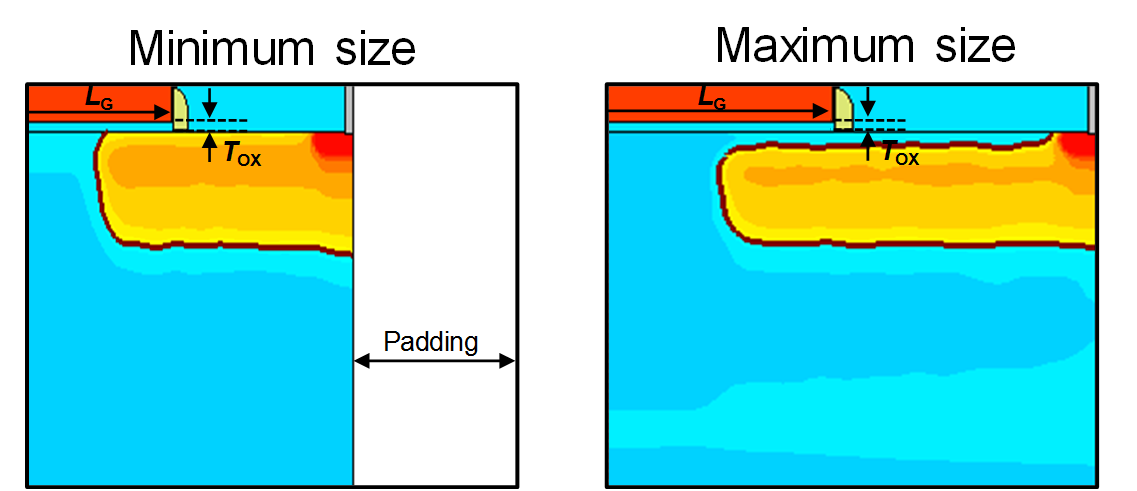}
   \caption{
   \textbf{Description of semiconductor datasets.}
   } \label{fig:7}
\end{figure}

\paragraph{Dataset Description.} We briefly introduce semiconductor process dataset, sampled by Technology Computer-Aided Design (TCAD, \cite{synopsys2009sentaurus}). The input features are 17 dimensions and consist of two types; the first is the features related to structure, and the other is related to the profile. Figure \ref{fig:7} presents the result of process simulator, explaining that the size of the doping profile depends on input features. Since structure-related features such as $L_G$ and $T_{OX}$ make the image size varying, we put padding area to maintain the fixed size. The profile-related features, the lower part of the image, also affect the color, which means when the yellow part shrinks, the blue part expands, or vice versa. All output images are added to padding to meet the images
320x320 size. 

\paragraph{Experiment Details.} Our experiment procedures are as follows: i) we trained the baseline models with 1,000 source data, ii) the number of target samples varied in size, iii) we assessed the performance of the methods with 51 target samples that domain engineers consider importantly.

\begin{figure}[h]
\begin{center}
\end{center}
   \centering
   \includegraphics[width=0.8\textwidth]{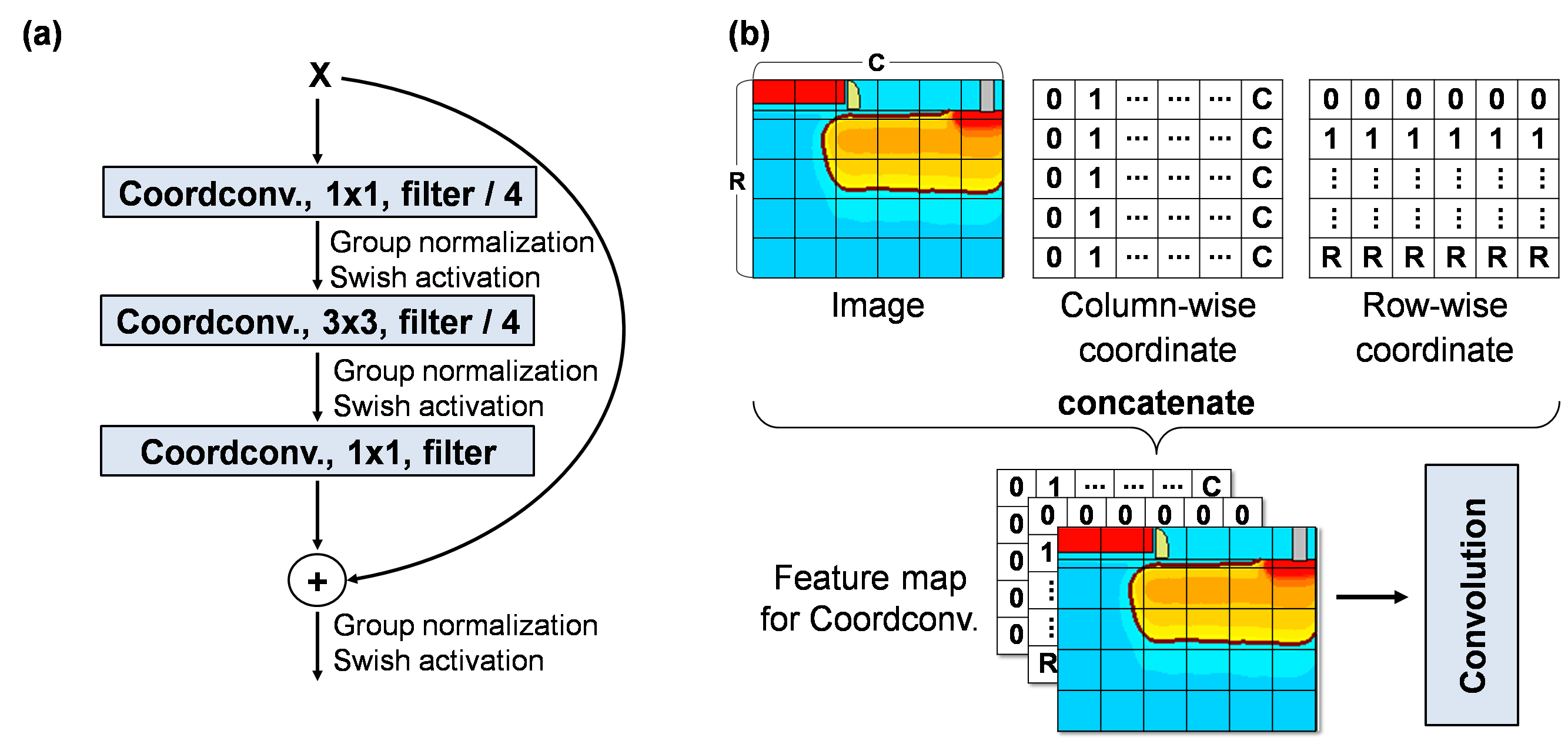}
   \caption{
   \textbf{Description of the baseline model on semiconductor dataset.} (a) Illustration of components of the residual block. (b) Description of coordinate convolution.
   } \label{fig:8}
\end{figure}

\paragraph{Architectural detail.} We used RTT model \cite{rtt, rtt2}. The model is based on residual blocks (\cite{resnet}), which consist of coordinate convolution (\cite{coordconv}), swish activation (\cite{swish}), and group normalization (\cite{wu2018group}) as depicted in Figure \ref{fig:8}. The model is trained for 10,000 epochs with a batch size of 128, where Adam optimizer is performed with the step size $10^{-4}$. For PAC-Net, we set pruning ratio is 0.8 and $\lambda = 0.01$.

\end{document}